\pdfoutput=1
\documentclass[runningheads]{llncs}

\usepackage{amsmath}
\usepackage{makeidx}  % allows for indexgeneration
\usepackage[misc]{ifsym} % For the envelope symbol for correspondant author
\usepackage[pdftex]{graphicx}
\usepackage{multirow}  % For table 
\usepackage{booktabs}  % For table
\usepackage{cite}

\begin{document}
\title{Parallel Capsule Networks for Classification of White Blood Cells}
\author{Juan~P.~Vigueras-Guill\'{e}n\inst{1}  \and   % \orcidID{0000-0002-5676-8217}   % index{Vigueras-Guillén, Juan P.}
	    Arijit~Patra\inst{2}     \and                % \orcidID{0000-0002-4447-6030}   % index{Patra, Arijit}
	    Ola~Engkvist\inst{3,4}     \and                % \orcidID{0000-0003-4970-6461}   % index{Engkvist, Ola}
	    Frank~Seeliger\inst{1}   }                   % \orcidID{0000-0003-0565-951X}   % index{Seeliger, Frank}
\authorrunning{Vigueras-Guill\'{e}n et al.}
%\institute{Anonymous}
\institute{CVRM Safety, Clinical Pharmacology and Safety Science, BioPharmaceuticals R\&D, AstraZeneca, Gothenburg, Sweden;\\  \and
	Digitisation and AI, Clinical Pharmacology and Safety Science, BioPharmaceuticals R\&D, AstraZeneca, Cambridge, United Kingdom;\\  \and
	Molecular AI, Discovery Science, BioPharmaceuticals R\&D, AstraZeneca, Gothenburg, Sweden;\\ \and
	Department of Computer Science \& Engineering, Chalmers University of Technology, Gothenburg, Sweden;\\
	\email{JuanPedro.ViguerasGuillen@astrazeneca.com}}
\maketitle              % typeset the header of the contribution
\begin{abstract}
Capsule Networks (CapsNets) is a machine learning architecture proposed to overcome some of the shortcomings of convolutional neural networks (CNNs). However, CapsNets have mainly outperformed CNNs in datasets where images are small and/or the objects to identify have minimal background noise. In this work, we present a new architecture, parallel CapsNets, which exploits the concept of branching the network to isolate certain capsules, allowing each branch to identify different entities. We applied our concept to the two current types of CapsNet architectures, studying the performance for networks with different layers of capsules. We tested our design in a public, highly unbalanced dataset of acute myeloid leukaemia images (15 classes). Our experiments showed that conventional CapsNets show similar performance than our baseline CNN (ResNeXt-50) but depict instability problems. In contrast, parallel CapsNets can outperform ResNeXt-50, is more stable, and shows better rotational invariance than both, conventional CapsNets and ResNeXt-50.

\keywords{CapsNets \and dynamic routing \and ResNeXt \and leukocytes.\footnote{This work was accepted to the International Conference MICCAI 2021.}}
\end{abstract}
\section{Introduction}

In the last decade, Convolutional Neural Networks (CNNs)  have shown remarkable performance for a wide range of computer vision tasks \cite{Krizhevsky2012, He2016, Huang2017}. However, CNNs have many drawbacks, such as the inability to learn viewpoint invariant representations and the need for large amount of training data. Capsule Networks (CapsNets) \cite{Sabour2017, Hinton2018} is a new neural network architecture that tackles those shortcomings by using capsules. A capsule is a group of neurons (depicted as a vector) whose output represents the various perspectives of an entity, such as pose, texture, scale, or the relative relationship between the entity and its parts. 

This technique has immense potential in the medical field, such as in cell classification, where (i) different types of cells are classified depending on the hierarchical relationship of the cell and its parts (shape of the nucleus, texture of the cytoplasm, presence of subcellular organelles), and where (ii) rotational invariance is crucial. However, CapsNets require a large number of parameters when the network is enlarged, and they have mainly shown promising performance for small images and/or with barely any background noise.

In this work, we present the concept of CapsNets parallelization, where parts of the network are subdivided in branches to isolate capsules, helping the network to (i) identify different entities in different branches, and (ii) avoid instability problems when capsule layers are enlarged. This concept is applied in both types of current CapsNets \cite{Sabour2017, Hinton2018}. We also propose a variation to the Sabour et al.'s CapsNets \cite{Sabour2017}; our proposal entails to lose the spatial information in the first layer of capsules, forcing the middle layer of capsules to encode whole entities. We also show how, against general assumption, conventional CapsNets do not seem to perform proficiently as more capsule layers are added, and they are not more robust than CNNs for small datasets.

\subsection{Capsule Networks}

\subsubsection{DR-CapsNets.}
Sabour et al. \cite{Sabour2017} proposed the first CapsNet based on dynamic routing (DR), with one CNN and two capsule layers, to solve the MNIST dataset (images of 28$\times$28 px). Their first layer of capsules, Primary-Caps, took the output of a convolution (6$\times$6$\times$256) and considered that every 8 elements along the feature axis would represent a capsule instantiation, thereby creating 32 capsules, each one evaluated in a grid of 6$\times$6. The latter simply entails that --in the subsequent steps-- the weights ($\textbf{W}_{ij}$) that multiply a lower-layer capsule $i$ to produce the next-layer capsule $j$ are shared between the capsules of the grid. Furthermore, they proposed that the module of a capsule vector should represent a probability (with range 0--1) and thus they defined a squashing function 
\begin{equation}
\textbf{v}_j = \frac{\mid\mid\textbf{s}_j\mid\mid^{2} }{1 + \mid\mid\textbf{s}_j\mid\mid^{2}} \frac{\textbf{s}_j}{\mid\textbf{s}_j\mid},  \ \ \ \ \   \textbf{s}_j = \sum_i c_{ij} \hat{\textbf{u}}_{j\mid i},  \ \ \ \ \ \hat{\textbf{u}}_{j\mid i} = \textbf{W}_{ij}\textbf{u}_i
\end{equation}
where $\textbf{v}_j$ and $\textbf{s}_j$ are the squashed and non-squashed capsule $j$, respectively, and the capsule $\textbf{s}_j$ is computed as indicated above for all (except the first) layer of capsules, where $c_{ij}$ are the coefficients obtained by the dynamic routing and $\textbf{u}_i$ are the squashed capsules from the lower layer. Briefly, the dynamic routing aims to determine how close the predicted vectors $\hat{\textbf{u}}_{j\mid i}$ are to the mean predicted vector $\textbf{v}_j$ (by using the scalar product), giving a higher $c_{ij}$ to those closer. They also defined a specific loss function, named \lq margin loss\rq. Further details in \cite{Sabour2017}.
% The network also included a decoder (a 3 neural network layers) to reconstruct the original image, used as a regularizer. 

\subsubsection{EM-CapsNets.}
Hinton et al. \cite{Hinton2018} proposed a different network (with one CNN and four capsule layers) to solve the smallNORB dataset (images of 96$\times$96 px). They extended the concept of CapsNets in the following ways: (i) capsules were depicted as matrices instead of vectors; (ii) a routing based on the Expectation-Maximization (EM) algorithm was proposed, where the matching between capsules is done by considering that a higher-layer capsule represents a Gaussian and the lower-layer capsules are data-points (further details in \cite{Hinton2018}); (iii) convolutional capsules were presented, where a higher-layer capsule is computed only based on the neighbouring lower-layer capsules (K in Fig.~\ref{fig1}); (iv) a technique called `Coordinate Addition' was proposed (applied only in the last convolutional capsule), which adds the position of the capsule to its vote matrix in order to keep the spatial information; (v) a new loss function, spread loss, was defined. 
% This network did not include a decoder.

\subsection{Related Work}

Many publications have used CapsNets to perform tasks in medical images. The majority simply applied Sabour et al.'s network \cite{Sabour2017} to small patches to perform tasks such as detection of diabetic retinopathy (fundus images) and mitosis (histology images, H\&E) \cite{Jimenez2018}, classification of breast cancer \cite{Iesmastas2018, Anupama2019} and colorectal tissue (H\&E) \cite{Nguyen2021}, detection of liver lesions (CT, adding attention gates to the network) \cite{Hoogi2019}, and blood vessel segmentation (fundus, with an inception block \cite{Szegedy2015} as CNN) \cite{Kromm2020}. Others used Sabour et al.'s network in larger images by adding more CNN blocks (glaucoma detection, OCT \cite{Gaddipati2019}) or heavily downsampling the images (brain tumour classification, MRI \cite{Afshary2018}). Only a few explored the use of a 3-capsule-layer DR-CapsNet for the classification of white cells \cite{Liu2019} and detection of cell apoptosis \cite{Mobiny2019}, and an interesting use of DR within a CNN architecture has been used for classification of thoracic disease (CT) \cite{Shen2018}. Special attention goes to LaLonde et al., who has developed CapsNets with several capsule layers to perform polyp classification \cite{LaLonde2020} and expanded the concept of CapsNets to perform segmentation (CT images of the lungs) \cite{LaLonde2018}. Hinton et al.'s architecture \cite{Hinton2018} has seldom been applied in the literature.

\section{Methods}

\subsection{Parallel Capsule Networks}

\begin{figure}[!t]
	\centering
	\includegraphics[width=0.95\textwidth]{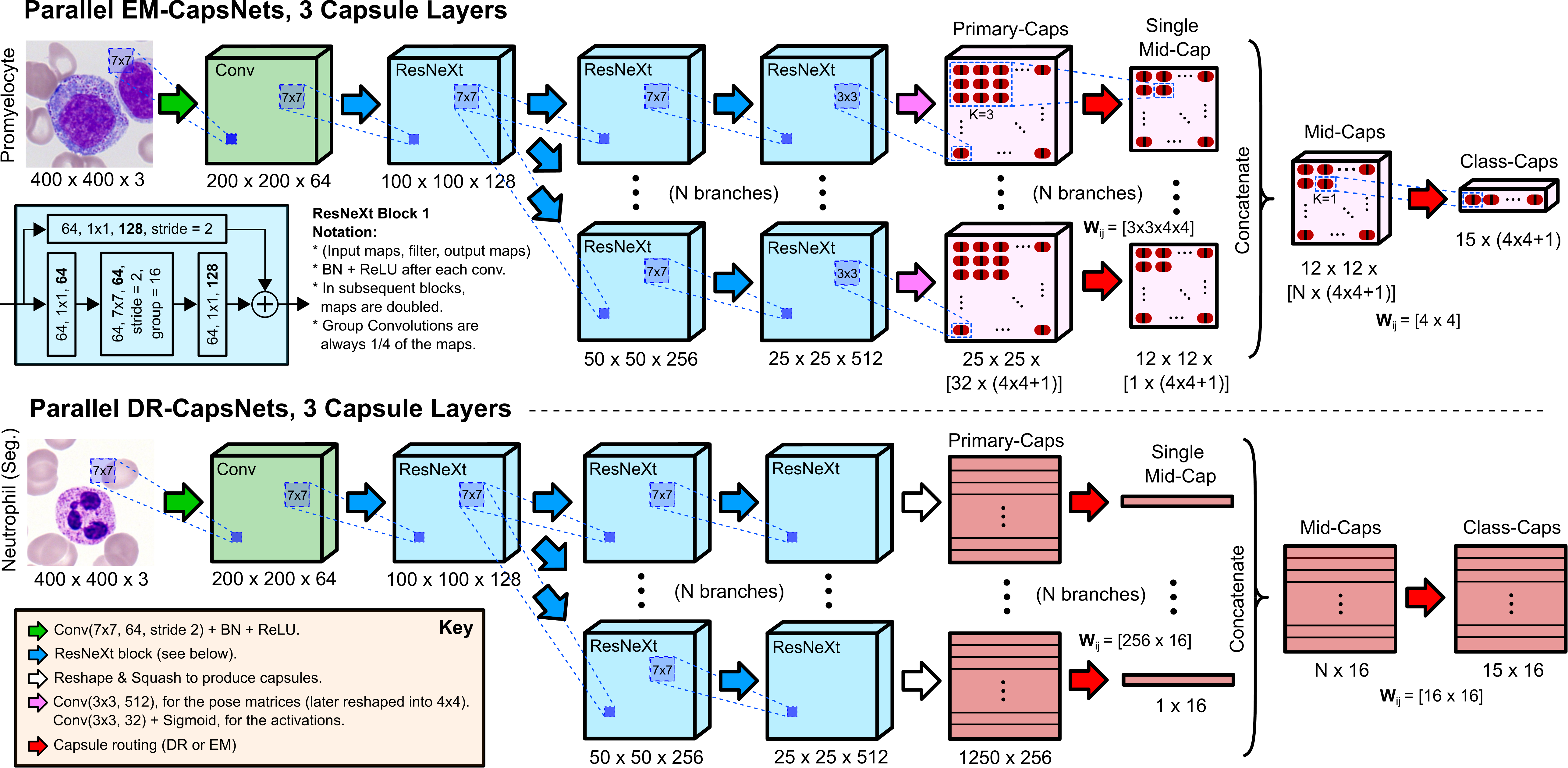}
	\caption{The proposed Parallel-CapsNets in two types of capsule architectures: EM (top and middle), and DR (bottom). An arrow represents an action, whereas a block represents the output of the action. \textit{Zoom for details}. } \label{fig1}
\end{figure}

Parallelization can be applied in several ways, depending where to start and merge the branches. In this work, we focus in studying networks with 3 capsule layers (3-CapsLayer). These contain Primary-, Mid-, and Class-Caps, and the parallelization is performed by creating a unique set of CNNs and Primary-Caps for each Mid-Cap, which are then concatenated to be routed to the Class-Caps (Fig.~\ref{fig1}, top). Several CNN-blocks precede the capsule section, allowing the image to be reduced to a suitable size for the capsules. The first CNNs, which generate basic features, are common to all branches. Our chosen CNN is ResNeXt \cite{Xie2017}. 

Since there is only one Mid-Cap per branch, there is no routing between Primary-Caps and Mid-Caps, which means that the algorithms only need to find the most appropriate transformation matrix $\textbf{W}_{i|j=1}$ per Primary-Cap. This allows each branch to focus in generating specialized features (either in the CNNs or Primary-Caps) that are suitable for the Mid-Cap.

For comparative purposes, networks with 2 and 4 capsule layers were also tested. The 2-CapsLayer does not include Mid-Caps, and thus the different sets of Primary-Caps are concatenated before routing to the Class-Caps. The 4-CapsLayer performs the merging before the Class-Caps.

\subsubsection{Parallel EM-CapsNets, 3 capsule layers.}
Our proposed network (Fig.~\ref{fig1}, top) applies parallelization to Hinton et al.'s architecture \cite{Hinton2018}. All capsules have a pose matrix of 4$\times$4 and an activation value. In a branch, there are 32 Primary-Caps and 1 Mid-Cap. The window (K) used in the convolutional capsules is 3$\times$3 for the Mid-Caps and 1$\times$1 for the Class-Caps, with a stride of 2 for the Mid-Caps. Coordinate addition is not employed.

\subsubsection{Parallel DR-CapsNets, 3 capsule layers.}
Our DR network (Fig.~\ref{fig1}, bottom) introduces two changes to Sabour et al.'s network \cite{Sabour2017}: the weights $\textbf{W}_{ij}$ are not shared among the capsules of the grid, and the Primary-Caps are 256 elements long (two capsules per grid-point). The former entails that the spatial information is lost in the first layer, forcing the Mid-Caps to encode the whole existence of an entity. Furthermore, since all Primary-Caps are directed to a single Mid-Cap, the algorithm can only discard a useless Primary-Cap by setting its $\textbf{W}_{i|j=1}$ to zero and, thus, not sharing weights can help to such discard. Both, the Mid-Caps and Class-Caps, have a size of 16 elements.  

\subsubsection{Parallel CapsNets, 2 capsule layers.}
In the EM network, the number of Primary-Caps are evenly distributed among the different branches, and the last CNN-block is duplicated to allow another stride. In the DR network, the length of the Primary-Caps is reduced based on the number of branches.

\subsection{Data, baseline networks, implementation details, and metrics}

\subsubsection{Data Description.}
We evaluate the proposed networks on a public dataset of white blood cells (leukocytes), which were present in patients with acute myeloid leukaemia (AML), a blood-type cancer that leads to the overproduction of abnormal leukocytes, published by Matek et al. \cite{Matek2019} (Laboratory of Leukemia Diagnostics at Munich University Hospital, Germany). The dataset contains 18,365 single-cell images of 400$\times$400 pixels from 15 highly-unbalanced classes (see Table~\ref{table_02}). This dataset is particularly interesting because (i) leukocytes show hierarchical structures (the nucleus might depict a prominent nucleolus or be formed by different segments, and cytoplasm might show different textures), (ii) it is highly unbalanced, and (ii) many red cells appear in the background (noise).

\subsubsection{Baseline.}
The network ResNeXt-50 \cite{Xie2017} was chosen as baseline. This network was also used by the dataset's authors \cite{Matek2019}, but we achieved a higher performance with the same network and setup: +6\% in PRE and SEN. DenseNets \cite{Huang2017} and ResNets \cite{He2016} were also tested, but they provided significantly inferior performance. An adapted single block of ResNeXt was chosen for the CNN-blocks in CapsNets, which also gave better results than other alternatives. We also tested the non-parallel versions of both architectures (EM and DR).

\subsubsection{Implementation Details.}
All networks were implemented in Tensorflow 2.2 on a single NVIDIA V100 GPU with 32GB of memory. In order to determine the most appropriate number of branches in the networks, each class in the dataset was subdivided into 5 folds, using 4 for training and 1 for validation. Once established the best branching, a 5-fold cross-validation (CV) was performed on the whole dataset for a final comparison (specifically, we use the same aforementioned 5 folds in the CV setup). The batches contained one example of each class (15 images), randomly shuffling the order within the batch to avoid bias. Data augmentation was performed by flipping the images up-down and left-right and by rotating the images (0$^{\circ}$--180$^{\circ}$). For the CNN baseline, the loss function was categorical cross-entropy. For the CapsNets, we used the loss functions suggested by their original authors (margin loss and spread loss). Nadam optimizer \cite{Dozat2016} was used, with a learning rate of 0.001. We defined an epoch as 500 iterations, and we trained for 700 epochs (no early stop), which required from 6 to 20 days to train, depending on the number of layers.

\subsubsection{Evaluation Metrics.}
To quantity the performance, we used the weighted (\textit{WAcc}) and non-weighted categorical accuracy (\textit{Acc}),
\begin{equation}
\text{WAcc} = \sum_{i} \dfrac{\text{TP}^i + \text{TN}^i}{\text{TP}^i + \text{TN}^i + \text{FP}^i + \text{FN}^i},  \ \ \text{Acc} = \dfrac{\text{TP} + \text{TN}}{\text{TP} + \text{TN} + \text{FP} + \text{FN}} ,
\end{equation}
reported as a percentage, where $i$ is the class, TP is true positives, TN is true negatives, FP is false positives, and FN is false negatives. We also defined an agreement metric (\textit{Agr}) that measures the percentage of images that were classified as the same type in all eight basic orientations (flipped up-down and left-right), regardless of whether the classification was correct. A higher agreement should suggest better rotational invariance. For the final architectures, we estimate the sensitivity, SEN = TP/(TP+FN), and precision, PRE = TP/(TP+FP).

\section{Results}

\subsubsection{Non-parallel architectures.}
In the DR-CapsNets, the 2-layer type performed rather poorly (Table~\ref{table_01}), the 3-layer increased performance but diverged with more than two Mid-Caps (Fig.~\ref{fig2}-B), and the 4-layer was unable to converge. Indeed, our proposed modification in DR networks (losing spatial information in the Primary-Caps) is not suitable for networks with only two layers. Regarding the EM network, it allowed for deeper architectures, but performance decreased with the addition of layers and/or Mid-Caps (Fig.~\ref{fig2}). Building a 4-layer EM network was feasible, but performance was poor in all possible setups (\textit{WAcc} in the range of 60--65). This suggests that, contrary to popular belief, the original EM-CapsNet does not perform as expected when the capsule section is increased in layers. EM-CapsNets were also highly sensitive to the position where capsule strides were placed, not being advised to place it in the Primary-Caps. Other instability problems were observed: EM networks were unable to converge when either (i) the image was only reduced to 50x50 before entering the capsule section (regardless if further striding was performed in the capsule section), or (ii) the Primary-Caps were increased from 32 to 64 (regardless the number of capsule layers). In summary, these findings suggest that both routing algorithms fail when a large number of capsules take part. Overall, only the 2-layer EM network (better \textit{WAcc}) and the 3-layer DR network (better \textit{Acc} and \textit{Agr}) were slightly better than ResNeXt-50 (Table~\ref{table_01}), and their number of parameters were notably lower (Table~\ref{table_01}).

\setlength{\tabcolsep}{3.8pt}
\begin{table}[!b]
	\caption{Categorical accuracy, weighted (\textit{WAcc}) and non-weighted (\textit{Acc}), and the agreement metric (\textit{Agr}) in the validation set for the best setups in each case (N is the number of branches), along with the number of parameters in the network (NPar). }
	\label{table_01}
	\centering
	\resizebox{\textwidth}{!}{
		\begin{tabular}{lcccccrcccccr}
			\toprule
			& & \multicolumn{5}{c}{\textbf{Non-Parallel}} & & \multicolumn{5}{c}{\textbf{Parallel}} \\
			\cline{3-7} \cline{9-13} \\[-9pt]
			\textbf{Network} & \textbf{Layers} & \textbf{N} & \textbf{WAcc} & \textbf{Acc} & \textbf{Agr} & \textbf{NPar} & & \textbf{N} & \textbf{WAcc} & \textbf{Acc} & \textbf{Agr} & \textbf{NPar} \\
			\midrule
			\multirow{2}{*}{DR-CapsNet} & 2  & - & 65.9 & 95.2 & 94.5\% & 77.3M &  % DONE!
			                                 & 4 & 59.4 & 95.4 & 95.1\% & 77.6M\\  % DONE!
			\cline{2-13} \\[-8pt]	
                            & 3 &  2 & 70.6 & 95.5 & 96.7\% & 11.3M &   % From Previous (done then!)
                                & 12 & 75.4 & \textbf{96.1} & \textbf{97.4}\% & 73.2M \\  % From Previous (done then!)
			\midrule										
			\multirow{2}{*}{EM-CapsNet} & 2 & - & \textbf{76.0} & 94.7 & 96.1\% & 3.6M &  % DONE
			                                & 2 & 75.4 & 93.9 & 94.6\% & 4.6M\\  % DONE
			\cline{2-13} \\[-8pt]	
                            & 3 &  2 & 69.4 & 93.0 & 94.9\% &  3.1M &     % DONE (To check the best N)
                                & 10 & 72.3 & 93.9 & 93.8\% & 30.1M \\    % DONE
			\midrule										
			ResNeXt-50         & -  & -    & 73.8 & 94.9 & 96.2\% & 23.3M & & - & - & - & - & -\\	% DONE
			\bottomrule
	\end{tabular}}
\end{table}
\setlength{\tabcolsep}{1.4pt}

\subsubsection{Parallel architectures.}
\begin{figure}[!t]
	\centering
	\includegraphics[width=\textwidth]{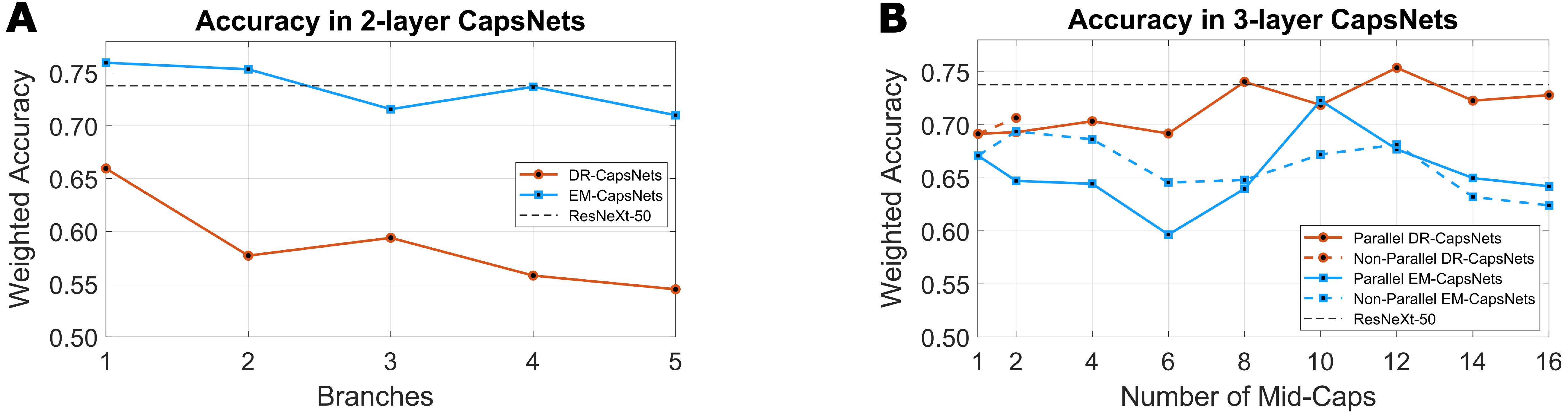}
	\caption{Weighted accuracy in the validation set for the (A) 2-CapsLayer and (B) 3-CapsLayer networks, based on the number of branches (N) or Mid-Caps. For the 2-layer CapsNets, a branch 1 indicates the non-parallel setup.} \label{fig2}
\end{figure}

As expected, the parallelization of 2-layer networks was rather detrimental: even though DR-CapsNets provided slightly better \textit{Acc} and \textit{Agr} in all cases, \textit{WAcc} was always lower (Fig.~\ref{fig2}-A), and EM-CapsNet provided lower results in all three metrics for all cases. In contrast, parallelizing 3-layer CapsNets was highly beneficial, particularly for our DR network (Fig.~\ref{fig2}-B), which did not depict convergence problems with the addition of branches. Regarding EM-CapsNets, the right number of branches yielded a peak in accuracy (Fig.~\ref{fig2}-B) (\textit{Acc} and \textit{Agr} were always higher than their non-parallel counterparts). In general, this agrees with our hypothesis that parallelization allows each branch to detect an entity independently (in its Mid-Cap) without affecting the remaining Mid-Caps. The drawback is the high increase in parameters. Specifically, it increased 3M parameters per each new branch in the EM network (only 0.01M per each Mid-Cap in the non-parallel version) and 6M in the DR network (5M in the non-parallel version). However, the increase in computational time was not excessive (a 0--50\% more). Overall, DR-CapsNet outperformed ResNeXt-50 with the right selection of branches (much higher \textit{Acc} and \textit{Agr}, Table~\ref{table_01}), but EM-CapsNet was still most proficient with only 2-layers and no branching (none of the EM nets --parallel or not-- outperformed ResNeXt-50 in \textit{Acc} and \textit{Agr}).

\subsubsection{Size of the capsules.}
The original CapsNets \cite{Sabour2017, Hinton2018} employed 16 elements to encode a capsule, but it was not discussed the relevance of that size or whether capsules from different layers should have different sizes. For this, we experimented with several sizes (9, 16, and 25, and their equivalent matrices 3$\times$3, 4$\times$4, 5$\times$5) in all possible combinations, and we found that (i) a size of 16 is the most optimal among our options, (ii) it is preferable --but not crucial-- that all layers have the same capsule size, (iii) a smaller capsule (9) worked surprisingly well as long as all layers employ it, and (iv) a larger size (25) started overfitting.

\subsubsection{Performance for small training data.}
To evaluate this, we split the dataset in the following way: up to $x$ images per class were used to create 5 folds, using 4 folds for training and 1 for testing, placing the remaining images above $x$ in the test set (being $x=[10,25,50,100,250,500,1000]$). For the lowest $x$, classes were balanced, becoming more unbalanced as $x$ increases. This was tested in the baseline ResNeXt-50 and the best CapsNets. Interestingly, our experiments depicted a similar behavior as $x$ increased for all cases, which contradicts the assumption that CapsNets outperform CNNs in smaller datasets.

\subsubsection{Overall benefits of our proposed networks.}
The non-parallel CapsNets showed convergence problems when a capsule layer became slightly large, which limits the number of entities to detect in the layer. However, there is no supportive evidence to believe that capsule layers should be fully connected between them. Indeed, a capsule (object) would only need to have connections to a reduced number of lower capsules (parts of the object), and different objects might be formed by completely different parts. Thus, branching would help to overcome the aforementioned limitation, although at the expense of a higher number of parameters. Moreover, the 3-layer CapsNets showed an interesting behaviour: the case with one single Mid-Cap was able to encode a generic white cell in just a capsule of 16 element that, subsequently, was transformed to 15 different classes. This highlights the power of capsules to encode entities and transformations. It also suggests that branching should be considered with heterogeneous datasets (groups of objects with morphological dissimilarities among the groups).

\setlength{\tabcolsep}{3.8pt}
\begin{table}[!t]
	\caption{Precision and sensitivity for the different classes in a 5-fold CV, for the ResNeXt-50 and the best CapsNets: the default 2-layer EM-CapsNet, and the parallel 3-layer DR-CapsNet with 12 branches. }
	\label{table_02}
	\centering
	\resizebox{\textwidth}{!}{
		\begin{tabular}{lrccccccccccc}
			\toprule
			& & & \multicolumn{2}{c}{\textbf{ResNeXt-50}} 
			& & \multicolumn{2}{c}{\textbf{EM-CapsNet}} 
			& & \multicolumn{2}{c}{\textbf{Par. DR-CapsNet}}\\
			\cline{4-5} \cline{7-8} \cline{10-11} \\[-9pt]
			\textbf{Type of cell} & \textbf{Images} 
			& & \textbf{Precision} & \textbf{Sensitivity} 
			& & \textbf{Precision} & \textbf{Sensitivity}  
			& & \textbf{Precision} & \textbf{Sensitivity}  \\
			\midrule
			Neutrophil (segm.)  & 8,484  & & 0.987 & 0.977 & & 0.992 & 0.958 & & 0.988 & 0.980 \\
			Neutrophil (band)   &   109  & & 0.330 & 0.523 & & 0.190 & 0.596 & & 0.350 & 0.520 \\
			Lymphocyte (typ.)   & 3,937  & & 0.988 & 0.924 & & 0.968 & 0.964 & & 0.991 & 0.930 \\
			Lymphocyte (atyp.)  &    11  & & 0.125 & 0.091 & & 0.667 & 0.182 & & 0.750 & 0.273 \\
			Monocyte            & 1,789  & & 0.913 & 0.908 & & 0.918 & 0.911 & & 0.930 & 0.917 \\
			Eosinophil          &   424  & & 0.964 & 0.943 & & 0.983 & 0.932 & & 0.955 & 0.962 \\
			Basophil            &    79  & & 0.732 & 0.658 & & 0.630 & 0.798 & & 0.658 & 0.705 \\
			Myeloblast          & 3,268  & & 0.904 & 0.975 & & 0.940 & 0.956 & & 0.934 & 0.988 \\
			Promyelocyte        &    70  & & 0.597 & 0.571 & & 0.662 & 0.643 & & 0.630 & 0.612 \\
			Promyelocyte (bil.) &    18  & & 0.393 & 0.611 & & 0.282 & 0.611 & & 0.465 & 0.611 \\
			Myelocyte           &    42  & & 0.579 & 0.524 & & 0.525 & 0.500 & & 0.611 & 0.524 \\
			Metamyelocyte       &    15  & & 0.316 & 0.400 & & 0.333 & 0.267 & & 0.380 & 0.466 \\
			Monoblast           &    26  & & 0.436 & 0.923 & & 0.414 & 0.923 & & 0.418 & 0.820 \\
			Erythroblast        &    78  & & 0.819 & 0.872 & & 0.892 & 0.846 & & 0.852 & 0.873 \\
			Smudge cell         &    15  & & 0.563 & 0.600 & & 0.643 & 0.600 & & 0.633 & 0.600 \\
			\midrule
			Total              & 18,365  & & 0.643 & 0.700 & & 0.669 & 0.712 & & 0.703 & 0.727 \\	
			\bottomrule
	\end{tabular}}
\end{table}
\setlength{\tabcolsep}{1.4pt}

\subsubsection{Final results}

Our proposed 3-layer parallel DR-CapsNet provided slightly better sensitivity and precision (Table~\ref{table_02}), but there was not a clear pattern (the detection of low-represented classes was not highly improved by CapsNets). Overall, we believe that three layers of capsules is appropriate for this dataset because of the morphological structure of white cells: the CNNs might denoise the image from background red cells while retaining the important features from the white cells, Primary-Caps might encode those basic features into capsules, Mid-Caps might then encode whole entities (nucleus, cytoplasm, or even generic whole cells), and Class-Caps might simply be the connection (and transformation) of different Mid-Caps entities. Our experiments also seemed to suggest that losing the spatial information in the layer previous to the merging is the most appropriate approach to exploit branching, but we could not test that hypothesis in the EM network due to lack of time. Many other experiments could also be tested to further improve the performance: branches with different sizes, merging some branches at different layers, etc.

\section{Conclusions}
Our work suggests that, for the classification of white cells, original CapsNets (i) do not generally outperform a well-established CNN (ResNeXt-50) unless it is a simple 2-layer network, therefore (ii) adding more capsule layers is usually detrimental, (iii) they are not more robust for small training data, (iv) they tend to be very sensitive to the tuning parameters, (v) they are unable to converge if a layer contains too many capsules, and (vi) their rotational encoding does not seem to be outstanding. In contrast, our proposed parallel DR-CapsNet seems to better learn the viewpoint invariant representations (highest \textit{Agr}), provides better accuracy (highest \textit{Acc}), and does not suffer from convergence problems.


\small
\bibliographystyle{splncs04} 
\bibliography{Paper1054_references_MICCAI}

\begin{thebibliography}{10}
\providecommand{\url}[1]{\texttt{#1}}
\providecommand{\urlprefix}{URL }
\providecommand{\doi}[1]{https://doi.org/#1}

\bibitem{Afshary2018}
Afshary, P., Mohammadiy, A., Plataniotis, K.N.: Brain tumor type classification
  via {Capsule Networks}. In: 25th IEEE International Conference on Image
  Processing (ICIP). pp. 3129--3133. Athens, Greece (2018)

\bibitem{Anupama2019}
Anupama, M.A., Sowmya, V., Soman, K.P.: Breast cancer classification using
  {Capsule Network} with preprocessed histology images. In: 2019 International
  Conference on Communication and Signal Processing (ICCSP). Chennai, India
  (2019)

\bibitem{Dozat2016}
Dozat, T.: Incorporating {Nesterov} momento into {Adam}. In: International
  Conference on Learning Representations Workshop (ICLRW). San Juan, Puerto
  Rico (2016)

\bibitem{Gaddipati2019}
Gaddipati, D.J., Desai, A., Sivaswamy, J., Vermeer, K.A.: Glaucoma assessment
  from {OCT} images using {Capsule Network}. In: 41st Annual International
  Conference of the IEEE Engineering in Medicine and Biology Society (EMBC).
  pp. 5581--5584. Berlin, Germany (2019)

\bibitem{He2016}
He, K., Zhang, X., Ren, S., Sun, J.: Deep residual learning for image
  recognition. In: 2016 IEEE Conference on Computer Vision and Pattern
  Recognition (CVPR). pp. 770--778. Las Vegas, NV, USA (2016)

\bibitem{Hinton2018}
Hinton, G.E., Sabour, S., Frosst, N.: Matrix capsules with {EM} routing. In:
  International Conference on Learning Representations (ICLR) (2018)

\bibitem{Hoogi2019}
Hoogi, A., Wilcox, B., Gupta, Y., Rubin, D.L.: Self-attention {Capsule
  Networks} for object classification. arXiv  \textbf{1904.12483} (2019)

\bibitem{Huang2017}
Huang, G., Liu, Z., {van der Maaten}, L., Weinberger, K.Q.: Densely connected
  convolutional networks. In: 2017 IEEE Conference on Computer Vision and
  Pattern Recognition (CVPR). pp. 2261--2269. Honolulu, HI, USA (2017)

\bibitem{Iesmastas2018}
Iesmantas, T., Alzbutas, R.: Convolutional {Capsule Network} for classification
  of breast cancer histology images. In: 15th International Conference on Image
  Analysis and Recognition (ICIAR). Póvoa de Varzim, Portugal (2018)

\bibitem{Jimenez2018}
Jiménez-Sánchez, A., Albarqouni, S., Mateus, D.: {Capsule Networks} against
  medical imaging data challenges. In: Intravascular Imaging and Computer
  Assisted Stenting and Large-Scale Annotation of Biomedical Data and Expert
  Label Synthesis. MICCAI Workshop. LABELS 2018, CVII 2018, STENT 2018. LNCS.
  vol. 11043 (2018)

\bibitem{Krizhevsky2012}
Krizhevsky, A., Sutskever, I., Hinton, G.E.: Imagenet classification with deep
  convolutional neural networks. In: Advances in neural information processing
  systems (NIPS). pp. 1097--1105 (2012)

\bibitem{Kromm2020}
Kromm, C., Rohr, K.: Inception {Capsule Network} for retinal blood vessel
  segmentation and centerline extraction. In: IEEE 17th International Symposium
  on Biomedical Imaging (ISBI). pp. 1223--1226. Iowa City, IA, USA (2020)

\bibitem{LaLonde2018}
LaLonde, R., Bagci, U.: Capsules for object segmentation. In: Medical Imaging
  with Deep Learning (MIDL) Conference. Amsterdam, The Netherlands (2018)

\bibitem{LaLonde2020}
LaLonde, R., Kandely, P., Spampinatox, C., Wallacey, M.B., Bagci, U.:
  Diagnosing colorectal polyps in the wild with {Capsule Networks}. In: 17th
  IEEE International Symposium on Biomedical Imaging (ISBI). pp. 1086--1090.
  Iowa City, IA, USA (2020)

\bibitem{Liu2019}
Liu, Y., Fu, Y., Chen, P.: {WBCaps}: a capsule architecture-based
  classification model designed for white blood cells identification. In: 41st
  Annual International Conference of the IEEE Engineering in Medicine and
  Biology Society (EMBC). pp. 7027--7030. Berlin, Germany (2019)

\bibitem{Matek2019}
Matek, C., Schwarz, S., Spiekermann, K., Marr, C.: Human-level recognition of
  blast cells in acute myeloid leukaemia with convolutional neural networks.
  Nature Machine Intelligence  \textbf{1},  538--544 (2019)

\bibitem{Mobiny2019}
Mobiny, A., Lu, H., Nguyen, H.V., Roysam, B., Varadarajan, N.: Automated
  classification of apoptosis in phase contrast microscopy using {Capsule
  Network}. IEEE Transactions on Medical Imaging  \textbf{31}(1),  1--10 (2019)

\bibitem{Nguyen2021}
Nguyen, H., Blank, A., Dawson, H.E., Lugli, A., Zlobec, I.: Classification of
  colorectal tissue images from high throughput tissue microarrays by ensemble
  deep learning methods. Nature Scientific Reports  \textbf{11:2371} (2021)

\bibitem{Sabour2017}
Sabour, S., Frosst, N., Hinton, G.E.: Dynamic routing between capsules. In:
  Proceedings of the 31st International Conference on Neural Information
  Processing Systems (NIPS). pp. 3859--3869 (2017)

\bibitem{Shen2018}
Shen, Y., Gao, M.: Dynamic routing on deep neural network for thoracic disease
  classification and sensitive area localization. In: 9th International
  Conference on Machine Learning in Medical Imaging (MLMI). Workshop. LNCS.
  vol. 11046, pp. 389--397. Granada, Spain (2018)

\bibitem{Szegedy2015}
Szegedy, C., Liu, W., Jia, Y., Sermanet, P., abd Ann~Arbor, S.R., Anguelov, D.,
  Erhan, D., Vanhoucke, V., Rabinovich, A.: Going deeper with convolutions. In:
  2015 IEEE Conference on Computer Vision and Pattern Recognition (CVPR).
  pp.~1--9. Boston, MA, USA (2015)

\bibitem{Xie2017}
Xie, S., Girshick, R., Doll{\'{a}}r, P., Tu, Z., He, K.: Aggregated residual
  transformations for deep neural networks. In: IEEE Conference on Computer
  Vision and Pattern Recognition (CVPR). pp. 5987--5995. Honolulu, HI, USA
  (2017)

\end{thebibliography}
\end{document}